# Implications of Deep Circuits in Improving Quality of Quantum Question Answering

Pragya Katyayan[1], Nisheeth Joshi[2]

[1,2]Department of Computer Science, Banasthali Vidyapith, Rajasthan.
pragya.katyayan@outlook.com; nisheeth.joshi@rediffmail.com

**Abstract:**

Question Answering (QA) has proved to be an arduous challenge in the area of natural language processing (NLP) and artificial intelligence (AI). Many attempts have been made to develop complete solutions for QA as well as improving significant sub-modules of the QA systems to improve overall performance through the course of time. Questions are the most important piece of QA, because knowing the question is equivalent to knowing what counts as an answer [1]. In this work, we have attempted to understand questions in a better way by using Quantum Machine Learning (QML). The properties of Quantum Computing (QC) have enabled classically intractable data processing. So, in this paper, we have performed question classification on questions from two classes of SelQA (Selection-based Question Answering) dataset using quantum-based classifier algorithms- quantum support vector machine (QSVM) and variational quantum classifier (VQC) from Qiskit (Quantum Information Science toolKIT) for Python. We perform classification with both classifiers in almost similar environments and study the effects of circuit depths while comparing the results of both classifiers. We also use these classification results with our own rule-based QA system and observe significant performance improvement. Hence, this experiment has helped in improving the quality of QA in general.

# 1 Introduction

NLP is a branch of AI and is arguably the most complex and challenging of all other areas. It has seen some crucial advances in AI. At the same time, QC has grown consistently in terms of hardware and algorithms that are classically intractable even with reasonable number of resources. This provides extensive opportunities for AI and specifically NLP to group with QC and find solutions to its most severe challenges of all times. Scientists have already proved quantum advantages for NLP tasks including algorithmic speedups and enhanced way to encode complex linguistic structures. QA is an AI-complete problem and we need to explore new dimensions to address it. QC and QML are such areas with the potential of being helpful in solving such problems.

The combination of QC and ML has the potential of changing how we look at previously unsolvable problems. It presents four fusion choices of both worlds as shown in Figure 1. The approaches are combinations of classical data processed with classical algorithms (cc), classical data processed with quantum algorithms (cq), quantum data processed with classical algorithms (qc) and quantum data processed with quantum algorithms (qq). Since, the repositories of classical data are immense and powerful set of processing capabilities are need-of-the-hour, we have considered processing classical data using quantum algorithms (i.e., cq). Hence, using quantum-enhanced machine learning (or QML) we have tried to utilize quantum algorithms and computers to tap in the extraordinary depths of information processing of classical data. Quantum algorithms are known to see data quite differently than classical ones [2]. If a quantum information processor can produce statistical patterns that are difficult for classical computers, then possibly, they can recognize better patterns in data than classical computers [3].

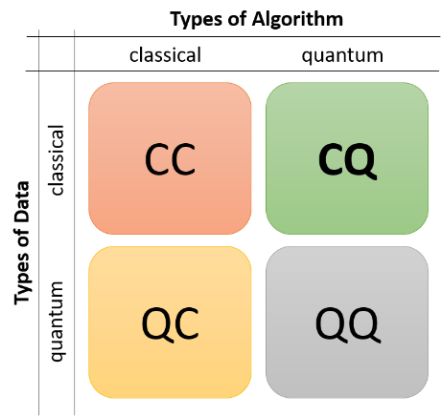

**Fig 1.** Four approaches to combine classical and quantum worlds out of which CQ is the most exciting and feasible given the availability of classical data today [2].

The most common yet most crucial tasks in text processing these days is classification. Every year, this task gathers significant amount of research work produced by researchers across the world. With this paper we attempt question classification over SelQA dataset that has questions from 10 classes. Question-Answering (QA) is a well-known challenge in the field of NLP and AI. It has three necessary steps: query processing, information retrieval, and response extraction. The query processing step plays a vital role in the whole process because knowing the question is equivalent to knowing what counts as an answer [1]. For instance, we can grab the domain of question [4] and its category [5] through query processing. Question classification provides the domain of possible answers along with type of answer desired by the question. Example:

**Question:** How many Despicable Me movies are there?
**Domain:** Movies
**Category:** Count

These are just two of many attributes possessed by a question. Such information is extracted by using the question classification method. This can be done using two approaches: rule-based [6] or statistics-based (i.e., ML) [5, 7-12]. Since, ML techniques have given some extraordinary results with specific features-set [5, 11], its every researcher's first choice when it comes to tasks like question classification. Studies have shown SVM as the best classifier in this case [8,9,11].

We propose the use of QML for question classification. Why do we need QML? The answer to this question actually lies in the Hilbert space. Quantum computation creates exponential number of basis states according to the availability of qubits in the system. For instance, if the quantum computer has 5 qubits, then 25 basis states can be produced by the computation. As the number of qubits increase, we get much more states to work with. The impact of this phenomenon can be realized by the fact that if we have 275 qubits, then the number of states we can represent through them is 2275, which is more the number of particles in the observable universe. Quantum processing gives us more power to represent possible states than classical processing ever can.

Quantum information processing works on qubits that are analogous to classical bits. However, these qubits can exist in the state of superposition that means, as a contrast to the classical bit, qubits can be 0 and 1 at the same time. Following the Dirac notation, we can say, the state of the qubit can be described as $|\psi\rangle = \alpha |0\rangle + \beta |1\rangle$, where $\alpha$ and $\beta$ are the complex numbers representing amplitudes of classical states $|0\rangle$ and $|1\rangle$. These complex numbers are subject to normalization condition such that $|\alpha|^2 + |\beta|^2 = 1$. When the state $|\psi\rangle$ is measured, the observation is either $|0\rangle$ or $|1\rangle$ with probability $|\alpha|^2$ or $|\beta|^2$ [13].

Figure 2 describes a rough framework of classical machine learning. In case of ML, we always initiate with data, which is provided to a model that gives us a prediction. We can score the prediction through a cost function and estimate how to update the model's parameters based on gradient-based techniques. In QML, we think of replacing this model, or perhaps some parts of this model with quantum computation that can be executed on a quantum computer while giving us some form of quantum advantage. This would process the classical data on a quantum device by looking at this data from quantum perspective. The involvement of quantum properties would help to find out more patterns in data than classical models.

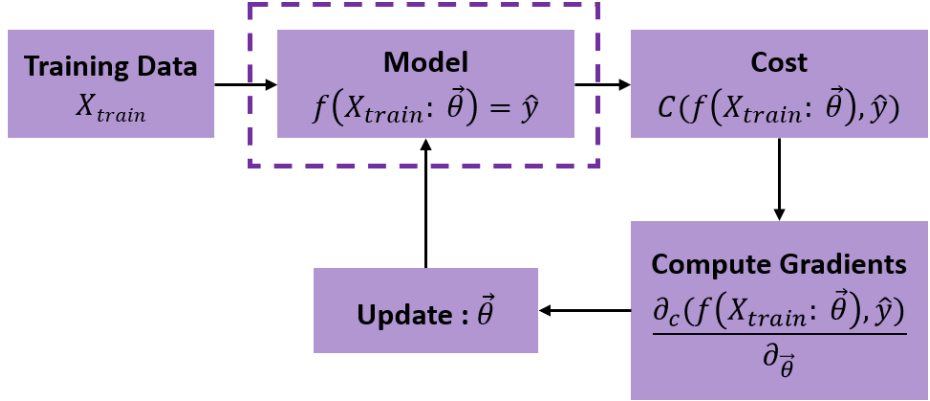

**Fig 2.** General framework of Machine Learning.

Studies in QML [3, 13-19] have shown its unique performance with different types of problems and different algorithms. The rise of QML started back in 1995 with the investigation in quantum-based neural network modeling. Then, in the early 2000s, researchers discussed statistical learning theory in quantum settings, which didn't get much attention. Around 2009, a workshop series in quantum computation took off where publications on quantum learning algorithms such as quantum associative memory, QBoost, etc. were observed. Around 2013, the term 'quantum machine learning' came into use, and since then, it became popular amongst researchers [2]. QML has many methods and algorithms to support different learning tasks. Under the umbrella of kernel classification methods lies the Quantum Support Vector Machine (QSVM) classifier, which has quite a good reputation in both classical and quantum worlds. In this work, we will use QSVM to classify questions and will analyze its performance.

Question classification is an important step in QA. In this paper, we attempt to improve the quality of QA, by better classification of questions. We perform two major tasks for this. First, we propose binary classification using two quantum classifiers- VQC and QSVM. We describe the experiments and analyze the results of these classifiers. We also present the repercussions of various combinations of hyperparameters for the quantum algorithms and the classification results for each combination with special emphasis on circuit-depth. Secondly, we also show the performance enhancement of our Rule-based Question-Answering (RBQA) system [36] after using the classification results from the first task as a feature.

The main contributions made through this paper are- implementation of quantum binary classification for real-life questions dataset with analysis of how the classifiers behave while classifying questions; practical details of QSVM and VQC configuration; insight about implications of circuit depth on classification; and proof of performance improvement of a QA system with the classification results.

The paper is arranged in the following way: section 2 consists of relevant literature reviews of work done in question classification and quantum machine learning; section 3 explains the dataset used for the experiment; section 4 elaborates on the features extracted; section 5, 6 and 7 explain important concepts and methodology of the research work done; section 8 talks about the experimental setup and section 9 gives the results and analysis. Last, but not the least, section 10 concludes the work throws some light on the possible future works.

## 2  Literature Review

### 2.1 Question classification

Hermjakob (2001) [7] has shown the requirement of semantically rich parse trees for better question classification using ML techniques. They have shown that question parsing improves the quality of question classification significantly. Zhang and Lee (2003) [8] have approached the problem of question classification with ML techniques trained on Bag of Words and Bag of n-grams features. They have observed the performances of Nearest Neighbors, Naïve Bayes, Decision Tree, Sparse Network of Winnows (SNoW), and SVM classifiers. In their paper, they have explained question classification and compared SVM with other ML algorithms and tree kernel and its calculation. Hacioglu and Ward (2003) [9] have taken up the ML approach for question classification to replace their previous rule-based classifier. In their paper, they have described the classifier. Metzler and Croft (2005) [10] have empirically shown that question classification can be done more efficiently by using statistical techniques. They have claimed that rule-based question classification can be too specific at times and can take a lot of effort to be crafted, while machine learning (statistical) techniques can prove suitable as it gets excellent performance with almost negligible efforts.

Huang, Thint & Qin (2008) [11] have emphasized the importance of features for better classification of questions. Instead of taking a rich feature space, they have opted for a compact yet effective feature set. Silva et al. (2011) [6] have identified question classification as an essential step for question-answering. They have presented a rule-based question classifier which finds the question headword and then maps it to its target category by using WordNet. They used WordNet to map headwords to categories of questions. They have also provided an empirical analysis of each feature's contribution to the best results. Li et al. (2017) [12] have adapted semi-supervised learning to do question classification. They have used a combination of classifiers, i.e., ensemble classifiers. They have compared the ensemble technique with a single classification technique. Liu et al. (2018) [5] claim that syntactic feature selection for question classification is computationally very costly. So, they have proposed a hybrid approach for semantic and lexical feature extraction. Gennaro et al. (2020) [20] have worked on classifying the intent of a question by using LSTM. They used Glove word embeddings to grab the semantic features of questions.

## 2.2 Quantum machine learning

Abramsky (2004) [25] has reported on the high-level methods for quantum computation. He has observed that the current tools are very low-level because there were loads of necessary computations, but the significance of high-level methods was not highlighted. Coecke (2004) [26] has reported on the logic of entanglement by exposing, with a theorem, the capabilities of the information flow of pure-bipartite entanglement. He has used this theorem to re-design and analyze popular protocols and has demonstrated the production of new ones. He has explained the extension of these results to the multipartite case. Aimeur et al. (2006) [13] have investigated the collaboration of machine learning and quantum information processing. Their approach is to define new learning tasks related to ML in a quantum mechanical world. Schuld et al. (2015) [14] have given a systematic overview of QML along with the technical aspects and approaches. They observed that the researchers worldwide had taken the use of computational costs as well as for translation od stochastic methods in quantum theory logic. Cai et al. (2015) [21] have observed the challenge of increasing data and the problems faced by classical computers in managing it. They report on the first entanglement-based two, four, and eight-dimensional vector classification in different clusters. Zeng & Coecke (2016) [22] have proposed the applications of quantum algorithms to NLP. They have observed the implementational challenges of the CSC model due to a shortage of classical computational resources. They have shown methods of resolving this issue using quantum computing. Makarov et al. (2017) [23] have given a brief overview of quantum logic concerning natural language processing. They have discussed the representation of sentences in quantum logic. They describe sentence representation, similarity analysis, quantum logic in diagrams as well as evaluation on NLP problems. Biamonte et al. (2017) [3] have reviewed recent quantum techniques including quantum speedups, classical and quantum machine learning, quantum principal component analysis, HHL algorithm, QSVM, and kernel methods, reading classical and quantum data in quantum machines and quantum deep learning.

Grant et al. (2018) [14] have demonstrated that more expressive circuits than the hierarchical structure can be used to achieve better accuracy and are capable of classifying highly entangled quantum states. They have explained in detail the effects of noise on classifiers' performance and deployment of classifiers on a quantum computer. Schuld et al. (2018) [24] have pointed out the need for quantum technologies that need fewer qubits and quantum gates and are error-proof. Such an approach is possible through variational circuits, which is very much beneficial for ML as these circuits learn the gates parameters. Their paper explains the circuit-centric classifier, its architecture along with optimization techniques used and graphical representation of gates. They have used hybrid gradient descent for training. They have provided a comprehensive analysis of the results achieved with the experimentation. Ciliberto et al. (2018) [15] presented a review on QML with a classical point of view. They have discussed possibilities for both classical and quantum experts. They have focused mainly on the limitations of quantum algorithms and where they stand in comparison to their classical competitions. They have also discussed why quantum is supposed to be a better resource for learning problems. Cardenaz-Lopez et al. (2018) [27] have proposed a protocol for performing quantum reinforcement learning. This protocol doesn't require coherent feedback during the learning process. Hence, it can be implemented in various types of quantum systems. Shukla et al. (2018) [28] have performed quantum process tomography of all the gates used in IBM processors and have computer gate-error to check the feasibility of complex quantum operations. They have also compared the quality of these gate with those built using other technologies. They have analyzed with this experiment that technological improvement would be required for achieving the scalability needed for quantum operations that are complex.

Kusumoto et al. (2019) [16] have exploited the complex dynamics of solid-state nuclear magnetic resonance to enhance machine learning. They have proposed to use Hamiltonian evolution based on inputs to map them to feature space. Havlicek et al. (2019) [18] have proposed two novel methods that they have implemented on a superconducting processor. In the first method, the quantum variational classifier was built on a variational quantum circuit and similarly classifies the training dataset as the classical SVM. In the second method, they have estimated kernel function and optimized the classifier. They developed artificial data that could be easily and completely be separated by the feature maps. This experiment was completed in two steps- first, the classifier was trained and optimized, and in second, the classifier labels the unlabeled data. Even in the presence of noise, their approach was able to touch a 100% success rate. Tacchino et al. (2019) [19] have observed that implementations of artificial neural networks in today's world are hindered by the growing computational complexity, which is a must for training MLP. So, they have introduced an algorithm based on quantum information that implements a binary-valued perceptron on a quantum computer. Mishra et al. (2019) [29] have used deep learning and supervised learning to hone quantum techniques by proposing a quantum neural network for cancer detection.

Meichanetzidis et al. (2020) [30] performed QA on NISQ device. In their model, the sentences were instantiated as parametrized quantum circuits with meanings encoded as quantum states. They took care of the grammatical structure explicitly by hard-wiring it as entanglement operations. This made this approach NISQ-friendly. Guo (2021) [31] introduced a density matrix which modelled a sentence-level attention mechanism. They developed a BiLSTM model based on word-level attention of weak measurement along with sentence-level attention for density matrix. They used this model to perform QA using WikiQA datasets. Song et al. (2021) [32] performed rigorous experiments on QA models to check if the relationship between weights of linguistic units and the outputs given by the model were uncertain. They observed that there were few ways where higher weights correlated with the model with a great impact, but usually the data didn't correspond well. This meant that the weight and model prediction were independent of each other and it was possible that different weight distributions produced similar results. Correia et al. (2022) [33] developed a theoretical pipeline to perform sentence disambiguation and question-answering which took advantage of quantum features. For the disambiguation task, their contraction scheme dealt with phrases that were syntactically ambiguous. For QA, they extracted a query representation which had all possible answers in equal superposition. Then they implemented Grover's search algorithm to find correct answer. Zhao et al. (2022) [34] found a balance between a model's performance and interpretability while proposed a quantum attention-based language model. Density matrix was used in quantum attention mechanism. The language model was applied in the answer selection module of a typical QA task to achieve effective results.

## 3 Dataset

The experiments for this research work were performed on a gold-standard benchmark dataset generated by Jurczyk, Zhai and Choi (2016) [19], which is popularly known as the Selection-based Question Answering (SelQA) dataset. This dataset has been developed from 486 Wikipedia articles as dumped in August 2014. The corpus has items from 10 most prevalent domains, i.e., Arts, Country, Food, Historical Events, Movies, Music, Science, Sports, Travel, and TV. The original data taken from the websites was broken down in smaller chunks during preprocessing. Each article was segmented into sections with the help of section boundaries available in the raw data. Further, every section was broken into sentences using NLP4J toolkit. The corpus has annotated question-answering examples on the above-mentioned topics. Each instance in the dataset consists of 7 different attributes, out of which the 'question' and its 'type' (i.e., domain) were extracted to make a custom corpus for the question classification task. The dataset was already divided into training data with 5529 sentences and testing data with 1590 sentences. The distribution of each class in the dataset is described in table 1. Data annotation was done through four sequential tasks on Amazon Mechanical Turk.

**Table 1.** Dataset distribution for each class in SelQA train and test set.

| Sl. No. | Classes in SelQA | Questions in Training dataset | Questions in Testing dataset | Total |
|---|---|---|---|---|
| 1. | Art | 467 | 135 | 601 |
| 2. | Country | 618 | 178 | 791 |
| 3. | Food | 509 | 147 | 652 |
| 4. | **Historical Events** | 571 | 164 | **730** |
| 5. | Movies | 574 | 164 | 735 |
| 6. | Music | 541 | 155 | 677 |
| 7. | **Science** | 621 | 179 | **795** |
| 8. | Sport | 585 | 168 | 741 |
| 9. | Travel | 573 | 165 | 734 |
| 10. | TV | 470 | 135 | 604 |
|  | **Total:** | **5529** | **1590** | **7060** |

Since, we have attempted binary classification using quantum classifiers, we have randomly chosen two classes- 'Historical Events' and 'Science' for the experiment. The classes have total 1525 (i.e., 730 and 795, respectively) unique questions. We have balanced the dataset by removing 65 datapoints from science class. Final dataset on which experiments were conducted had 730 questions from each domain, i.e., total 1460 questions out of which 80% datapoints were taken for training and rest 20% for testing purpose.

## 4 Feature Selection

Since we wish to perform binary classification, we take up two questions of two classes from the SelQA dataset, one from 'Historical Events' class and another from 'Science' domain. The two domains are quite different from each other in terms of keywords, which plays a crucial role in accomplishing classification. We extracted eleven distinct features from the questions. Different experiments were performed with different groups of features. For instance, some experiments were performed with four features, some with five, some with seven features while the rest were performed with complete 11 features. The features are mentioned in table 2, along with features considered in different groups. They are further elaborated in the subsequent sections.

**Table 2.** Features extracted for SelQA dataset

| Sl. No. | Feature Name | Feature groups used in experiments | | | |
|---|---|---|---|---|---|
| | | 4 | 5 | 7 | 11 |
| 1. | Content Words | ✓ | ✓ | ✓ | ✓ |
| 2. | Non-content Words | ✓ | ✓ | ✓ | ✓ |
| 3. | Question keywords | ✓ | ✓ | ✓ | ✓ |
| 4. | Wh-words | ✓ | ✓ | ✓ | ✓ |
| 5. | Nouns | | ✓ | ✓ | ✓ |
| 6. | Verb count | | | ✓ | ✓ |
| 7. | 1-gram Probabilities | | | ✓ | ✓ |
| 8. | 2-gram Probabilities | | | | ✓ |
| 9. | 3-gram Probabilities | | | | ✓ |
| 10. | 4-gram Probabilities | | | | ✓ |
| 11. | 5-gram Probabilities | | | | ✓ |

### *4.1 Content and Non-content words*

Content words are responsible for the important information in a sentence. Non-content words have semantic information where they facilitate anticipation of some feature of the words that follow. They are also called function words which help connect the content words. These are responsible for helping the model better understand the semantics and capture the context. We counted the content words and non-content words of all the questions as two important features. Content words comprised of Nouns,

Verbs, Adjectives and Adverbs and other words were counted as non-content words. Examples are given in table 3.

**Table 3.** Examples of content and non-content words count from questions

| Questions | Content words | Non content words |
|---|---|---|
| how many times was the who national fyrd called out between 1046 and 1065? | 6 | 9 |
| when did the germans begin using chlorine gas on the western front? | 8 | 5 |
| as a result of the napoleonic wars the british empire rose in power beginning the historical period known as what? | 11 | 10 |
| what is the natural satellite of earth? | 4 | 4 |
| what two creations of confirmation bias are under study with respect to astrological belief? | 8 | 7 |

## *4.2 Question Keywords*

We have removed the general English stop-words from the questions dataset and have used the remaining keywords as a feature. These keywords consist of significant words from the individual domains. These are markers that help the classifier in better classification. These keywords are converted into vectors with the help of Bag-of-Words. Examples are given in table 4.

**Table 4.** Examples of keywords extracted from questions

| Questions | Q_keywords |
|---|---|
| how many times was the who national fyrd called out between 1046 and 1065? | ['many', 'time', 'national', 'fyrd', 'call'] |
| when did the germans begin using chlorine gas on the western front? | ['german', 'begin', 'use', 'chlorine', 'gas', 'western', 'front'] |
| what is the natural satellite of earth? | ['natural', 'satellite', 'earth'] |
| what two creations of confirmation bias are under study with respect to astrological belief? | ['two', 'creation', 'confirmation', 'bias', 'study', 'respect', 'astrological', 'belief'] |
| what had faraday concluded based on his electrochemical experiments? | ['faraday', 'conclude', 'base', 'electrochemical', 'experiment'] |

## 4.3 Wh-words

Every question either has a wh-word (how, what, when, where, which, who) or it starts with a verb ending with a question mark (eg. Does he have a gold coin?). We have captured the presence of all the wh-words and have marked the others as 'NA'. This will help in categorizing wh-questions. Since Wh-words list is a closed set of six words, we have extracted only those. Examples are given in table 5.

**Table 5.** Examples of wh-words extracted from questions

| Questions | Wh-words |
|---|---|
| how many times was the who national fyrd called out between 1046 and 1065? | how who |
| when did the germans begin using chlorine gas on the western front? | when |
| what is the natural satellite of earth? | what |
| what two creations of confirmation bias are under study with respect to astrological belief? | what |
| what had faraday concluded based on his electrochemical experiments? | what |

## 4.4 Nouns

Nouns in any sentence speaks of the significant entities- people, places and things. By identifying Nouns of any domain, we can easily identify set of significant people, places and things of that particular domain. These are keywords which help in answer generation. Examples are given in table 6.

**Table 6.** Examples of nouns extracted from questions

| Questions | Nouns |
|---|---|
| how many times was the who national fyrd called out between 1046 and 1065? | times fyrd |
| when did the germans begin using chlorine gas on the western front? | germans chlorine gas front |
| what is the natural satellite of earth? | satellite earth |
| what two creations of confirmation bias are under study with respect to astrological belief? | creations confirmation bias study respect belief |
| what had faraday concluded based on his electrochemical experiments? | faraday experiments |

## 4.5 Verb count

The number of verbs occurring in a sentence displays what kind of domain it is. The count helps us catch the syntactic formation of the system and helps the model learn what the sentence is made up of. It helps in identifying the complexity of sentence. Examples are given in table 7.

**Table 7.** Examples of verb count for different questions

| Questions | Verb count |
| --- | --- |
| how many times was the who national fyrd called out between 1046 and 1065? | 2 |
| when did the germans begin using chlorine gas on the western front? | 2 |
| what is the natural satellite of earth? | 1 |
| what two creations of confirmation bias are under study with respect to astrological belief? | 0 |
| what had faraday concluded based on his electrochemical experiments? | 3 |

## 4.6 N-gram probabilities

N-grams (where N $\epsilon$ {1, 2, ..., 5}) of texts are extensively used in NLP tasks to mine textual data for useful information. These are set of co-occurring words within a particular word-window (e.g., 2-grams = 2-word window). Since we are not doing grammatical analysis, n-gram approach is a mechanism of doing the same with statistical learning. We used language modelling to find the occurrences of these n-grams and calculate probabilities of their presence throughout the corpus. We considered probability values up to 5-grams as separate features for our dataset.

## 5   Variational Models

Variational models are quantum circuit-based models which have certain parameters that we can tweak, train, and optimize. Figure 3 shows a variational circuit where $U$ is any set of unitary gates that construct the feature map. $U(x)$ has parameters x that can be tweaked to get desired results. Next block is $W(\theta)$ which is the model circuit that helps in accomplishing the classification task where $\theta$ is the parameter to tweak, train and optimize the circuit. $M$ is the measurement unit. We have used this variational model as a classifier for our classical data to accomplish question classification. The task is- to train a quantum circuit $W(\theta)$, on labelled samples of data from $U(x)$, in order to get predictions of labels for unseen data. The first step is to encode the classical data into quantum state so that the quantum circuit can understand it using a feature map. The next step will apply a variational model on the encoded data that will be trained as a classifier. In the third step we will measure the classifier circuit to extract labels and last, but not the least we will use optimization techniques to update the model parameters.

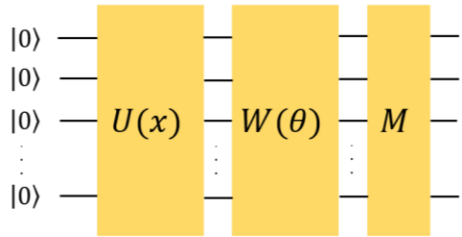

**Fig 3.** Representation of a variational model where the quantum circuit consists of parameterized gates that can be changed to get better results.

In the above diagram, $W(\theta)$ block is the classifier model. There are several circuits[1] available in Qiskit for such usage viz. $RealAmplitudes, EfficientSU2, TwoLocal, NLocal$ etc. For our experiment, we have taken $TwoLocal$ circuit provided by Qiskit's circuit library as the classifier model for training. It is a parametrized circuit with alternating rotation and entangled layers. The rotation layers are composed of single qubit gates which are applied on all qubits and the entanglement layer uses two-qubit gates to entangle the qubits. We can provide the combination of gates we wish to use for the rotation layer and entanglement. We used a fully entangled circuit to get the maximum quantum advantage possible.

## 6  Quantum Support Vector Machines

QSVMs were introduced by Havlicek et al. (2019) [17]. Quantum Kernel Estimation (QKE) implements SVM with a quantum kernel using quantum processing twice in the process. Firstly, kernel $K(x_i, x_j)$ is estimated on a Quantum device for each pair of training data $x_i, x_j \in T$. This kernel can later be used in Wolfe-dual SVM to detect the optimal hyperplane. Secondly, the quantum device is again used to estimate the kernel $K(x_i, s)$ for any new datapoint $s \in S$ and the support vectors from the set $x_i \in T$ obtained from the optimization. This is enough to construct the full SVM classifier. Quantum kernel support vector classification algorithm has the following steps:

1. Build the train and test quantum kernel matrices.
   a) For each pair of datapoints in the training dataset $x_i, x_j$, apply the feature map and measure the transition probability: $K_{ij} = |\langle 0|U^\dagger_{\Phi(x_j)}U_{\Phi(x_i)}|0\rangle|^2$.
   b) For each training datapoint $x_i$ and testing point $y_j$, apply the feature map and measure the transition probability: $K_{ij} = |\langle 0|U^\dagger_{\Phi(y_j)}U_{\Phi(x_i)}|0\rangle|^2$.
2. Use the train and test quantum kernel matrices in a classical support vector machine classification algorithm.

---

[1] https://qiskit.org/documentation/apidoc/circuit_library.html#n-local-circuits

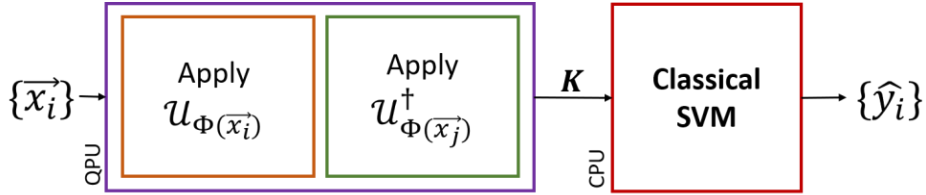

**Fig 4.** General architecture of quantum support vector machines as introduced in [17]

The quantum kernel should be hard to estimate classically as compared to their classical counterparts in order to tap into quantum advantage for quantum kernel machine algorithms [17]. We have encoded our datapoints using PauliFeatureMap with full entanglement (explained in the next section) which is hard to achieve classically and that dataset is supplied to the QPU for QKE. The processing and results are tough to simulate classically hence, our QSVM gets the quantum advantage over classical SVM.

## 7    Data Encoding

In this paper, we propose binary classification of classical data using two quantum classifiers. The classical processes of data encoding are followed during data preprocessing. However, classical data is not quantum readable and hence we need to map the classical data into quantum state space. For classical data, data encoding is usually done using single qubit rotations. The most efficient approach considered is to encode classical data in amplitudes of a superposition which means utilizing N qubits to encode 2N dimensional data vector. Although, this method is less efficient in terms of space but is very efficient in terms of time because it involves only single-qubit rotations [35]. Here data vectors are re-scaled in an element-wise manner to make them lie between $\left[0, \frac{\pi}{2}\right]$. Next step is to encode every vector element from the qubit using the following equation (eq. 1):

$$\psi_n^d = \cos(x_n^d)|0\rangle + \sin(x_n^d)|1\rangle \qquad (1)$$

Where, classical dataset for binary classification is a set $D = \{(x^d, y^d)\}_{d=1}^{D}$, with $x^d \in \mathbb{R}^N$ as N-dimensional input vectors and $y^d \in \{0,1\}$ are the corresponding data labels. The final data vector is written as $\psi^d = \bigotimes_{n=1}^{N} \psi_n^d$ and is ready to be used in quantum algorithm [35].

There are several ways to encode classical data to quantum state viz. basis encoding, amplitude encoding, angle encoding and higher order encoding. Since, we have complex dataset to work with, we go for higher order encoding using quantum feature maps. The choice of quantum feature maps depends on the type of data we have, however, there is no standard rules to decide which feature map suits which type of data. If the feature map is hard to simulate classically, that gives us quantum advantage. There are several feature maps provided by the Qiskit library which are capable of higher order encoding viz. *ZFeatureMap*, *ZZFeatureMap* and *PauliFeatureMap*. Since, we are using Pauli feature map to encode our classical data to quantum space, and it is proved to be hard to simulate classically [17], it gives us quantum advantage. Given the complex nature of our data, PauliFeatureMap seems the perfect choice to encode our data as it provides customisable combinations of Pauli Gates which could prove beneficial to map our data on the Hilbert space. We experimentally tried all three feature maps provided by the Qiskit library on small sample dataset carefully taken out from the original SelQA dataset and found PauliFeatureMap to be giving the best results. PauliFeatureMap when customized with Pauli gates: $P_0 = X, P_1 = Y, P_2 = ZZ$; can be explained by the following equation (eq. 2):

$$\mathcal{U}_{\Phi(x)} = \left( \exp\left(i \sum_{jk} \Phi_{\{j,k\}}(x) Z_j \otimes Z_k\right) \exp\left(i \sum_{j} \Phi_{\{j\}}(x) Y_j\right) \exp\left(i \sum_{j} \Phi_{\{j\}}(x) X_j\right) H^{\otimes n} \right)^d \qquad (2)$$

Each datapoint in our dataset has 11 features representing it, so total qubits we need would be 11. We kept the circuit short-depth with reps=1 and fully entangled.

# 8 Experimental Setup

Our dataset has labelled questions from two domains, historical events and science. We first clean and pre-process the dataset from any unwanted punctuations or notations and turn them all in lower-case. We then extract 11 features from the questions. Next, we normalize the features and split the dataset in the ratio of 80:20 where 80% of the dataset would be used for training the classifiers and 20% will be used as testing set. The dataset distribution is given in table 8. We encoded the classical data to quantum state using PauliFeatureMap. Further, we used two quantum classifiers to get classification results on the same dataset. First was QSVM and the other was VQC.

QSVM tries to detect a hyperplane in the input feature space that can separate the two classes. There are different kernel methods available that can be used to find out a hyperplane in high-dimensional input feature space without defining it. Kernels help to project the vectors in the feature space to a higher-dimensional vector space where a linear decision boundary separating two classes is possible. We have used Qiskit Aqua's QSVM module for this experiment. As we have discussed above, we chose PauliFeatureMap for this experiment with a short depth of 1, feature dimension 11 and fully-entangled to get quantum advantage. We ran the quantum instance on Qiskit's IBM Qasm Simulator which is a context-aware simulator and can simulate up to 32 qubits. Since, there is no IBM Q device with more than five qubits available on the cloud, we had to run our experiment on the Qasm simulator. We used a Linux-based system with Ubuntu 20.04 LTS operating system, Quadro RTX 5000 GPU to run this experiment and it took 62 hours to complete. The results are discussed in the next section.

**Table 8:** Dataset distribution across training and test set.

| Labels | Classes | Train | Test | Total |
|---|---|---|---|---|
| 0 | Historical Events | 534 | 146 | 730 |
| 1 | Science | 534 | 146 | 730 |
| | **Total:** | **1168** | **292** | **1460** |

VQC, on the other hand, takes another circuit as a model to train as a classifier. As we have discussed above, we have taken TwoLocal circuit of Qiskit library as the model with 'ry' and 'rz' as rotation layer gates and 'cz' as entanglement layer gate. We have used PauliFeatureMap with similar specifications as we used for QSVM. Finally, we have used COBYLA optimizer to optimize the results with maximum iterations of 100. We executed this experiment on IBM's Qasm simulator and the same hardware as QSVM experiment, and it took about 24 hours to complete. The results are discussed in the next section.

**Application of question classification on a pre-developed QA system:**

We developed a rule-based open-domain question answering (RBQA) system with hand-crafted rules for QA. The system was developed and tested on SQuAD 2.0 dataset [36]; however, the rules were capable of acting on any similar dataset. Figure 5 describes the working of RBQA system in a nutshell. We tested the RBQA system on questions from above-mentioned two classes of SelQA dataset. The system gave a satisfactory performance with ~65% accuracy. One notable nature of our system is that it displays exact answers only if it is able to fetch them, in all other cases the system displayed the most relevant sentence from the context, that might have the correct answer. Keeping this nature in mind, we applied these classification results as question-feature to our rule-based open-domain question answering (RBQA) system. With this additional feature the QA system gave more exact answers than earlier.

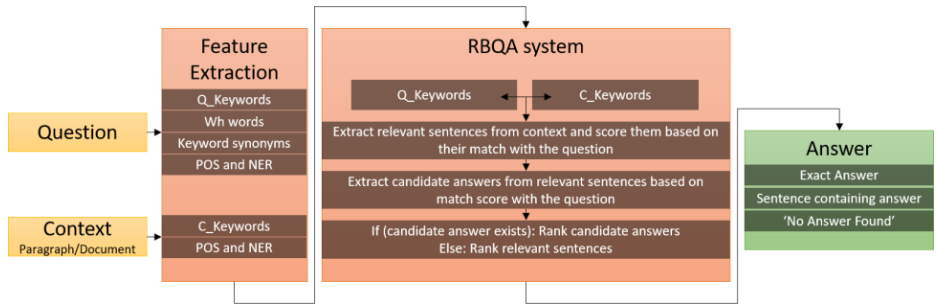

**Fig 5.** General working of the RBQA system developed by Katyayan & Joshi (2022) [36] explained pictorially.

The detailed results of all experiments are given in the next section.

## 9   Results and Analysis

Quantum machine learning has tremendous potential of solving problems which has been proved by researchers [17]. We chose to use its potential in solving a challenge from NLP domain- Question classification. We had SelQA dataset with questions from 10 different classes. Since we wish to accomplish binary classification only, we segregated two classes – Historical Events and Science from the dataset and balanced it. We extracted 11 features from the dataset and used two classifiers- QSVM and VQC to classify them. We kept the environment similar for both the classifiers, in order to get a comparative result. The results show that VQC is quite faster than QSVM. However, the testing accuracy of QSVM came higher than VQC. QSVM gave 60.61% accuracy while VQC gave 58.21%. This shows that QSVM classified the data more accurately than VQC. There were several challenges to this experiment including the unavailability of bigger IBMQ devices and more computing power. We took a considerably small dataset that took QSVM 60+ hours to process.

The results show a comparison between the QSVM classifier and VQC. These two experiments were executed on the same specifications (e.g., dataset, feature dimension, feature map, optimizers and other hyperparameters and hardware). The execution time of both the classifiers had a big gap. While VQC completed in 24 hours, QSVM took 62 hours to finish. A possible reason for this would be QSVM's dual processing of the QKE followed by hyperplane estimation while VQC runs just one classifier circuit. Given extra processing of QSVM, the extra time for execution is justified. The results of all the experiments run with both classifiers are presented in table 9 and 10.

**Table 9:** Results of various experiments run with QSVM (with change in hyperparameters)

| Exp no. | No. of Features | FM | Entanglement | FM Depth | FM Combination | Testing Accuracy |
|---|---|---|---|---|---|---|
| 1 | 2 | PauliFM | full | 1 | ['X', 'Y', 'ZZ'] | 55.98 |
| 2 | 2 | PauliFM | full | 2 | ['X', 'Y', 'ZZ'] | 50.71 |
| 3 | 4 | PauliFM | full | 1 | ['X', 'Y', 'ZZ'] | 53.58 |
| 4 | 4 | PauliFM | full | 2 | ['X', 'Y', 'ZZ'] | 50.63 |
| 5 | 5 | PauliFM | full | 1 | ['X', 'Y', 'ZZ'] | 49.50 |
| 6 | 7 | PauliFM | full | 1 | ['X', 'Y', 'ZZ'] | 53.11 |
| 7 | 7 | PauliFM | full | 2 | ['X', 'Y', 'ZZ'] | 48.85 |
| **8** | **11** | **PauliFM** | **full** | **1** | **['X', 'Y', 'ZZ']** | **60.61** |
| 9 | 11 | PauliFM | full | 2 | ['X', 'Y', 'ZZ'] | 54.21 |
| 10 | 11 | PauliFM | full | 3 | ['X', 'Y', 'ZZ'] | 53.94 |

**Table 10.** Results of various experiments run with VQC (with change in variables and hyperparameters)

| Exp no. | Features | FM | Entanglement | FM depth | FM combination | QC depth | Testing Accuracy |
|---|---|---|---|---|---|---|---|
| 1 | 2 | Pauli | full | 1 | ['X', 'Y', 'ZZ'] | 3 | 56.45 |
| 2 | 4 | Pauli | full | 1 | ['X', 'Y', 'ZZ'] | 4 | 62.2 |
| 3 | 4 | Pauli | full | 2 | ['X', 'Y', 'ZZ'] | 3 | 54.54 |
| 4 | 5 | Pauli | full | 1 | ['X', 'Y', 'ZZ'] | 1 | 53.77 |
| 5 | 5 | Pauli | full | 2 | ['X', 'Y', 'ZZ'] | 2 | 54.79 |
| 6 | 5 | Pauli | full | 3 | ['X', 'Y', 'ZZ'] | 3 | 55.47 |
| 7 | 7 | Pauli | full | 1 | ['X', 'Y', 'ZZ'] | 1 | 57.7 |

| Exp no. | Features | FM | Entanglement | FM depth | FM combination | QC depth | Testing Accuracy |
|---|---|---|---|---|---|---|---|
| 8 | 7 | Pauli | full | 2 | ['X', 'Y', 'ZZ'] | 2 | 51.44 |
| 9 | 7 | Pauli | full | 2 | ['ZZ' 'XY' 'ZZ'] | 2 | 46.88 |
| **10** | **11** | **Pauli** | **full** | **1** | **['X', 'Y, 'ZZ']** | **1** | **58.21** |
| 11 | 11 | Pauli | full | 2 | ['X', 'Y, 'ZZ'] | 2 | 53.76 |

In case of QSVM (table 9), we performed 10 different experiments with changes in hyperparameters. Subtle changes in hyperparameters lead to changes in testing accuracies. We started off with just 2 features and lowest feature map depth (i.e., 1) and we got a testing accuracy of 55.98%. Then we increased the depth to 2 and the accuracy fell to 50.11%. Next, we increased the number of features to 4 with depth 1, accuracy increased a little and came up to 53.58%. On increasing the depth to 2 the accuracy again fell to 50.63%. Next, we increased the features to 5 and attempted classification at depth 1, but the accuracy fell to 49.50%. So, we decided to add more features before trying out some more. On increasing the features to 7 at depth 1 accuracy rose to 53.11, but on increasing depth to 2 it again fell to 48.85%. Finally, we added 4 more features to the list and again trained the system on 11 features. With highest depth 3 the accuracy rose to 53.94%, on decreasing depth to 2- accuracy rose to 54.21% and with lowest possible depth 1 the accuracy rose to 60.61%.

In case of VQC, its circuit contains of both the feature map as well as a parametrized circuit. There can be numerous possibilities of depth value combinations along with endless combinations for Pauli gates in the Feature map circuit. So, to keep the study diverse as well as finite, we kept the Pauli gate combination set to the default of X+Y+ZZ and depth of both circuits are changed simultaneously to observe the results. Since, VQC performs well on short depth circuits [17], we stopped increasing depths if the accuracy value fell. We started observing with 2 features, FM depth 1 and circuit depth 3 and the accuracy we got was 56.45%. This accuracy was better than that of QSVM's first case with lowest depth. So, we increased the features to 4 and increased the circuit depth to 4. The testing accuracy jumped to 62.2%. We tried to check if the accuracy fluctuates on increasing FM depth and increased it to 2 while decreasing circuit depth to 3. Accuracy fell to 54.54%. Next, we increased the features to 5 and kept both the depths to the lowest value 1, the accuracy fell a little to 53.77%. We increased both depth values to 2 and got an accuracy of 54.79%. We again increased the depths to 3 each and got a slight accuracy rise of 55.47%. These were very slight changes in accuracy and so we again added couple features to the list and trained the system. With each depth values set to 1, we got accuracy of 57.7%; each depth values set to 2, we got a fall in accuracy value to 51.44%. Since the accuracy started falling with increased depths, we tried a different Pauli gates combination for the feature map on depth values 2 for each circuit, but the accuracy fell drastically to 46.88%. So, we understood that changing the Pauli gates combination in this case is probably not suitable, so we used the default combination of X+Y+ZZ gates. We increased the features to 11 and on lowest depths of both circuits we achieved an accuracy of 58.21%. On attempting for a higher depth value of 2 for both circuits, the accuracy fell to 53.76%.

The results in above tables show the change in accuracy as the number of features and circuit-depth were changed. Since, these algorithms are executed on noisy devices, short-depth circuits perform well as they are compliant with error-mitigation techniques that are capable of reducing decoherence effect [17]. In case of QSVM, when the number of features were increased gradually the accuracy started falling. However, with 11 features the accuracy reached 60% on a short depth circuit. We tried increasing the depth of feature map but that worsened the accuracy in every case, so we can say in case of question classification QSVM works best on short-depth circuits. We kept the feature map fully entangled to ensure quantum advantage. In case of VQC, the results get high with 4 features and deeper circuit, however, apart from that the results improve when circuit depth decreases and at maximum 11 features and shortest possible depth, the accuracy hits 58.21%. If we are to compare QSVM and VQC, then QSVM (exp-8) and VQC (exp-10) match similar hyperparameter values and hence the results are comparable. Table 11 shows few comparable results of both classifiers.

**Table 11.** Final results of accuracy given by QSVM and VQC for question classification task on same environment variables.

| Features | QSVM | Testing Accuracy (%) | VQC | | | Testing Accuracy (%) |
|---|---|---|---|---|---|---|
| | FM depth | | FM depth | QC depth | Total | |
| 4 | 1 | 53.58 | 1 | 4 | 5 | 62.2 |
| | 2 | 50.63 | 2 | 3 | 5 | 54.54 |
| 5 | 1 | 49.50 | 1 | 1 | 2 | 53.77 |
| 7 | 1 | 53.11 | 1 | 1 | 2 | 57.7 |
| | 2 | 48.85 | 2 | 2 | 4 | 51.44 |
| **11** | **1** | **60.61** | **1** | **1** | **2** | **58.21** |
| | 2 | 54.21 | 2 | 2 | 4 | 53.76 |

In the above table, it is clearly observed that since QSVM applies only one feature map circuit and VQC requires both feature map and a quantum circuit to work with, the circuit depth of QSVM will always be lower than VQC. QSVM consistently performs poorly than VQC under similar circumstances but with complete set of 11 features the testing accuracy reaches an all-time high value of 60.61%. Perhaps this particular set of features helped the classifier in finding a better hyperplane. At one instance, VQC surprised us with a 62.2% accuracy at the highest depth of 5. However, in every other case of VQC the accuracy value fell on increasing depth of the circuit. Also, as compared with QSVM, the testing accuracy in every other comparable case is higher, but with the complete feature set of 11 features, it fell a little behind with 58.21% testing accuracy. So, overall, we understand that for this particular case of question classification given the combination of features we are using, both QSVM and VQC performed well on short-depth circuits and feature set considered for experiments also affects the accuracy significantly. Stronger the feature, better the classification. Figures 6 and 7 display classifiers' performance pictorially. Figure 6 shows classifier-wise performance based on circuit depth. It can be observed easily that low circuit depths get high accuracy results. Figure 7 highlights the role of features on the increasing accuracy of both the classifiers.

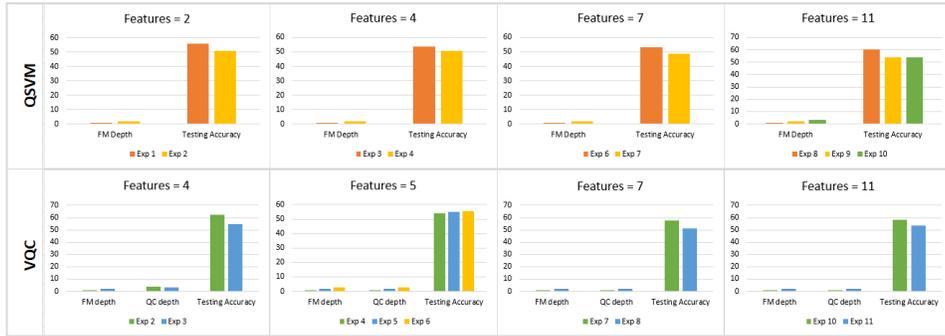

**Fig 6.** Graphs showing variations in testing accuracy on increasing circuit depth in different experiments (in case of QSVM, circuit depth = FM depth; in case of VQC, circuit depth = FM depth + QC depth)

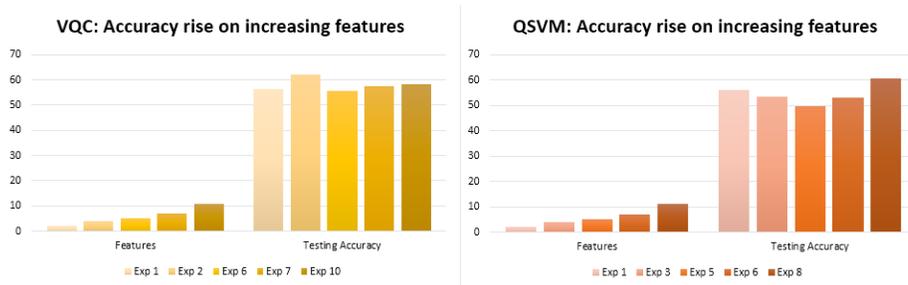

**Fig 7.** Graphs showing accuracy rise on increasing features for both VQC and QSVM.

**Performance of RBQA system while using QSVM's best classification results as features:**

The second task of our experiment was to apply the best classification results and study the implications on the accuracy of the QA system. First, we established a baseline by testing our RBQA system on the questions from SelQA dataset as defined in table 8. Our system gave 65% (i.e., 949/1460) correct answers including exact answers and relevant answer sentences. Out of the correctly answered 949 questions, 432 questions got exact answers while 517 got relevant sentences as answers. Next, we used QSVM's best classification result as feature-set for the questions that got correctly answered by our system. We observed that number of exact answers increased from 432 to 461. This happened because the new feature-set that was introduced, helped the system in finding relevant answers by defining the domain of the question. For instance, if system got confused between multiple candidate answers, the domain helped it in accurately selecting the correct answer. This proves that good quality question classification can help in improving the quality of QA.

## 10  Conclusion and Future Works

We have experimentally implemented binary question classification using quantum classifiers. We performed several experiments with different hypermetric settings. However, we kept few things constant just to keep the focus on circuit depth and avoid many complex permutations and combinations of all hyperparameters. We selected the best possible options and fixed them. For instance, we used PauliFeatureMaps with default Pauli gates settings in all the cases of both classifiers as they best represent complex data. The experiments show classification results at various depths. Observation of results showed us that in majority of the cases, both the classifiers perform well on short-depth circuits and QSVM slightly outperformed VQC while working under almost similar circumstances. Even on lowest depths, the classifiers' results vary according to different set of features. The performance falls initially on increasing features but when trained on complete 11 features, the testing accuracy rises. This realization enables us to conclude that strong features must be chosen for representing classical data. We had access to few quantum computers but none of them had enough qubits for our experiments. So, the results shown here are simulated using IBM's QASM simulator (supports up to 32 qubits). This also brings to light the vacancy of high-qubit capacity quantum computers for such experiments which involve higher number of features (and hence need more qubits to encode those features). The best classification results achieved by QSVM was used as additional feature in a QA system which proved to be a performance booster.

Future works may include training the classifiers on a bigger dataset and testing the classifiers' performance on a real quantum computer (NISQ devices). Also, if high-capacity quantum computers are made available in future, it would be interesting to note the performance of these classifiers on a real quantum computer.